\documentclass{article}

\usepackage[preprint]{neurips_2025}

\usepackage[utf8]{inputenc} 
\usepackage[T1]{fontenc}    
\usepackage{hyperref}       
\usepackage{url}            
\usepackage{booktabs}       
\usepackage{amsfonts}       
\usepackage{nicefrac}       
\usepackage{microtype}      
\usepackage{xcolor}         
\usepackage{multirow}
\usepackage{multicol}
\usepackage{amssymb}
\usepackage{enumitem}
\usepackage{amsmath}
\usepackage{graphicx}
\usepackage{bbding}
\usepackage{pifont}

\title{MLLM-Enhanced Face Forgery Detection: A Vision-Language Fusion Solution}

\author{
	Siran~Peng$^{1,2}$
	\quad Zipei~Wang$^{1,2}$ 
	\quad Li~Gao$^{3}$
	\quad Xiangyu~Zhu$^{1,2}$ \\
	\textbf{\quad Tianshuo~Zhang$^{2,1}$
		\quad Ajian~Liu$^{1,2}$
		\quad Haoyuan~Zhang$^{2,1}$
		\quad Zhen~Lei$^{1,2,4}$\textsuperscript{\dag}} \\
	{$^1$CASIA} \quad {$^2$UCAS} \quad {$^3$CMCC} \quad {$^4$CAIR, HKISI, CAS}\\
	\scriptsize{\texttt{\{pengsiran2023,wangzipei2023,xiangyu.zhu,ajian.liu,zhanghaoyuan2023,zhen.lei\}@ia.ac.cn}}\\
	\scriptsize{\texttt{gaolids@chinamobile.com, tianshuo.zhang@nlpr.ia.ac.cn}}\\
}

\begin{document}

\maketitle

\begin{abstract}
Reliable face forgery detection algorithms are crucial for countering the growing threat of deepfake-driven disinformation. Previous research has demonstrated the potential of Multimodal Large Language Models (MLLMs) in identifying manipulated faces. However, existing methods typically depend on either the Large Language Model (LLM) alone or an external detector to generate classification results, which often leads to sub-optimal integration of visual and textual modalities. In this paper, we propose VLF-FFD, a novel Vision-Language Fusion solution for MLLM-enhanced Face Forgery Detection. Our key contributions are twofold. First, we present EFF++, a frame-level, explainability-driven extension of the widely used FaceForensics++ (FF++) dataset. In EFF++, each manipulated video frame is paired with a textual annotation that describes both the forgery artifacts and the specific manipulation technique applied, enabling more effective and informative MLLM training. Second, we design a Vision-Language Fusion Network (VLF-Net) that promotes bidirectional interaction between visual and textual features, supported by a three-stage training pipeline to fully leverage its potential. VLF-FFD achieves state-of-the-art (SOTA) performance in both cross-dataset and intra-dataset evaluations, underscoring its exceptional effectiveness in face forgery detection.
\end{abstract}

\section{Introduction}
Recent breakthroughs in deepfake generation technology \cite{li2019faceshifter,Nirkin_2019_ICCV,9320155,Hsu_2022_CVPR,Shiohara_2023_ICCV} have sparked widespread interest, primarily because of their capacity to create hyper-realistic synthetic faces that closely mimic genuine human features. Although these advancements hold promise for applications in entertainment and creative industries, they also introduce significant risks, including enabling fraud, amplifying misinformation, and creating deceptive fake news. Consequently, there is a pressing demand for accurate and reliable methods to detect facial forgeries and counteract their potential harm.

\begin{figure}[t]
	\begin{center}
		\begin{minipage}{1\linewidth}
			{\includegraphics[width=1\linewidth]{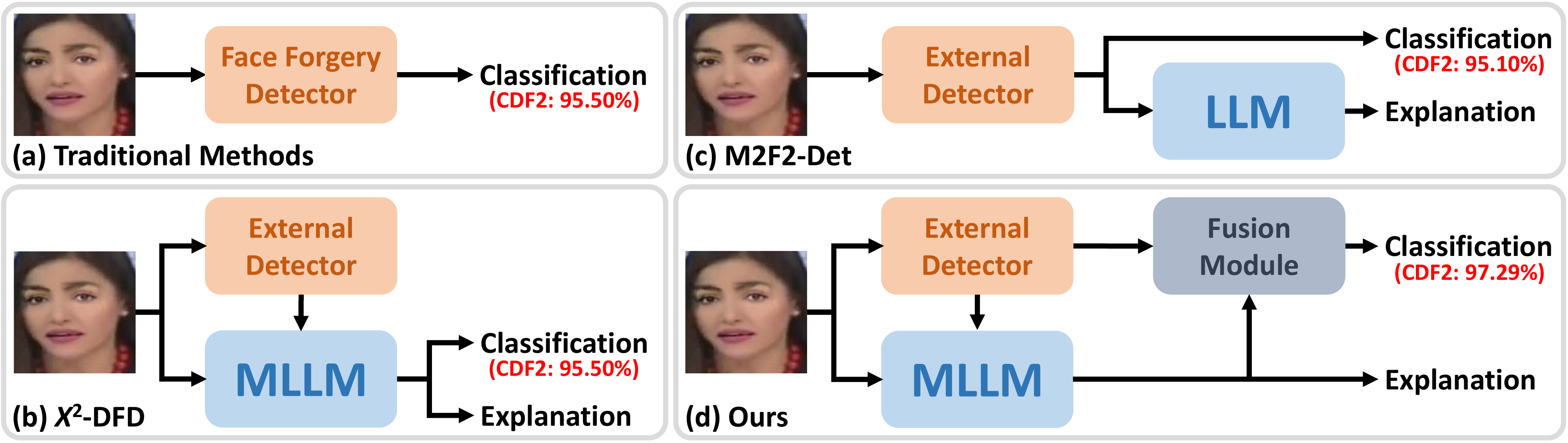}}
			\centering
		\end{minipage}
	\end{center}
	\caption{Comparative overview of face forgery detection frameworks evaluated on the Celeb-DeepFake-v2 (CDF2) benchmark \cite{Li_2020_CVPR}. \textbf{(a)} Traditional methods use face forgery detectors exclusively for classification, achieving a cross-dataset AUC of 95.50\%. \textbf{(b)} $\mathcal{X}^2$-DFD \cite{chen2024textit}, representing the first category of MLLM-based approaches, leverages an LLM (also MLLM) for both classification and explanation, also reaching 95.50\% AUC. \textbf{(c)} M2F2-Det \cite{guo2025rethinking}, belonging to the second category of MLLM-based methods, utilizes an external detector for classification and an LLM for explanation, achieving an AUC of 95.10\%. \textbf{(d)} The proposed VLF-FFD enables bidirectional interaction between visual and textual modalities, delivering superior performance with a 97.29\% AUC on CDF2. \label{hp}}
\end{figure}

The rapid development of Multimodal Large Language Models (MLLMs) has greatly improved the alignment between visual and textual modalities \cite{pmlr-v202-li23q,NEURIPS2023_6dcf277e,Liu_2024_CVPR,liu2024llava}. Recently, these models have been applied to face forgery detection, with an emphasis on generating explanatory insights. As shown in Figure~\ref{hp}, current MLLM-based methods typically adopt one of two strategies: (1) using the Large Language Model (LLM) to perform both classification and explanation \cite{shi2024shield,Jia_2024_CVPR,chen2024textit,yu2025unlocking}, or (2) incorporating an external detector for classification while utilizing the LLM for explanatory purposes \cite{guo2025rethinking}. However, both strategies rely on a single modality to deliver the classification results, which leads to sub-optimal integration of visual and textual information and, consequently, restricting overall performance.

To address this challenge, we propose VLF-FFD, a novel Vision-Language Fusion framework designed for MLLM-enhanced Face Forgery Detection. Our method consists of two main components. 
\textbf{First}, to facilitate more effective and informative MLLM training, we present EFF++, a frame-level, explainability-driven extension of the FaceForensics++ (FF++) dataset \cite{Rossler_2019_ICCV}, leveraging a commercial LLM \cite{team2023gemini}. Specifically, we introduce the Contrastive Forgery Artifacts Discovery (CFAD) method, which precisely identifies facial forgery clues by providing the commercial LLM with paired real and fake face images. Additionally, we expand the dataset by incorporating the Manipulation Technique Summarization (MTS) approach, which distills the manipulation strategies outlined in the original FF++ paper. As a result, for each manipulated video frame, EFF++ provides a paired textual annotation that describes both the forgery artifacts and the specific manipulation technique used.
\textbf{Second}, we design a Vision-Language Fusion Network (VLF-Net) that promotes bidirectional interaction between visual and textual features. Our network architecture comprises an external detector \cite{Liu_2022_CVPR}, an MLLM \cite{Liu_2024_CVPR}, and an innovative Vision-Language Fusion Module (VLFM). 
The external detector directly extracts visual features from the input image, while the MLLM processes both the input image and the visual feature map produced by the external detector to generate text embeddings and provide explanatory content. Finally, the VLFM integrates these visual and textual representations to generate the final classification result. 
To maximize the performance of VLF-Net, we employ a three-stage training pipeline: (1) training the external detector independently to ensure robust visual feature extraction, (2) freezing the external detector and training the MLLM to enhance textual representations, and (3) freezing both the external detector and the MLLM while training the VLFM to optimize the fusion process.
In conclusion, the \textbf{contributions} of this paper are as follows:
\begin{itemize}[leftmargin=12pt]
	\item We present EFF++, a frame-level, explainability-driven extension of the FF++ dataset. In EFF++, each manipulated video frame is paired with a textual annotation describing both the forgery artifacts and the manipulation technique, facilitating more informative training of the MLLM.
	\item We introduce VLF-Net, a novel architecture that integrates an external detector, an MLLM, and a fusion module to promote bidirectional interaction between visual and textual features. Additionally, we employ a three-stage training pipeline to maximize the potential of this network.
	\item The proposed VLF-FFD, consisting of the EFF++ dataset and the VLF-Net architecture, demonstrates state-of-the-art (SOTA) performance in both cross-dataset and intra-dataset evaluations, highlighting its exceptional effectiveness in the field of face forgery detection. Furthermore, comprehensive ablation studies validate the robustness and correctness of our design choices.
\end{itemize}

\section{Related Works}
\subsection{Traditional Face Forgery Detection}
Face forgery detection is traditionally framed as a binary classification problem, where a model is trained to assess whether an image or video is ``real'' or ``fake''. As outlined in \cite{pei2024deepfake}, these detection methods can be broadly divided into four categories: spatial-domain, time-domain, frequency-domain, and data-driven approaches.
Spatial-domain techniques identify facial forgeries by analyzing inherent spatial features within an image, such as color inconsistencies \cite{8803740}, saturation variations \cite{8803661}, and the presence of artifacts \cite{Zhao_2021_CVPR,Cao_2022_CVPR,Shiohara_2022_CVPR,9694644,cui2024forensics}. 
Time-domain methods, on the other hand, detect manipulations across temporal sequences by examining inter-frame inconsistencies, leveraging the entire video for analysis \cite{Haliassos_2021_CVPR,10.1145/3474085.3475508,gu2022delving,gu2022hierarchical,10054130,10478974,zhang2025learning}.
Frequency-domain techniques convert spatial or temporal data into the frequency domain using algorithms such as Fast Fourier Transform (FFT) \cite{tan2024frequency}, Discrete Cosine Transform (DCT) \cite{qian2020thinking,Li_2021_CVPR}, and Discrete Wavelet Transform (DWT) \cite{9447758,10.1145/3503161.3547832,10004978,peng2025wmamba}. These approaches excel at revealing subtle forgery traces that may not be visible in the spatial or temporal domains. 
Lastly, data-driven methods refine network architectures and optimize training strategies, fully harnessing the potential of available data to improve detection performance \cite{Zhao_2021_ICCV,hu2022finfer,Guo_2023_CVPR,Huang_2023_CVPR,Yan_2023_ICCV,Zhai_2023_ICCV,Yan_2024_CVPR}.
However, the approaches mentioned above only deliver classification results without offering any explanations for their decisions, which limits their interpretability.

\subsection{MLLM-based Face Forgery Detection}
\label{s22}
Vision and language represent two of the most powerful modalities in Artificial Intelligence (AI). Recent advancements in MLLMs, particularly the LLaVA series \cite{NEURIPS2023_6dcf277e,Liu_2024_CVPR,liu2024llava}, have significantly enhanced the alignment and integration of visual and textual data. These models are now being actively leveraged for face forgery detection, with a key emphasis on providing explanatory insights \cite{shi2024shield,Jia_2024_CVPR,huang2024ffaa,NEURIPS2024_059d2b91,xu2025identity}. Current MLLM-based detection methods can be roughly classified into two categories, depending on how they produce classification results. The first category utilizes the LLM to handle both classification and explanation tasks simultaneously. Notable examples include $\mathcal{X}^2$-DFD \cite{chen2024textit} and KFD \cite{yu2025unlocking}. In $\mathcal{X}^2$-DFD, the output score from an external detector is integrated into the prompt to fine-tune the LLM, enabling it to generate both classification results and explanations. Similarly, KFD transforms the outputs of the external detector into prompt embeddings, which fine-tune the LLM to produce the final classification and explanation. The second category incorporates an external detector for classification, while the LLM is employed to generate explanations. A prominent example in this category is M2F2-Det \cite{guo2025rethinking}, which uses a CLIP-based \cite{pmlr-v139-radford21a} detector to produce classification results and prompt embeddings. These embeddings then guide the LLM in generating explanations. However, methods in both categories rely on a single modality for final classification, leading to sub-optimal integration of visual and textual information, which ultimately limits the overall performance.

\begin{figure}[t]
	\begin{center}
		\begin{minipage}{1\linewidth}
			{\includegraphics[width=1\linewidth]{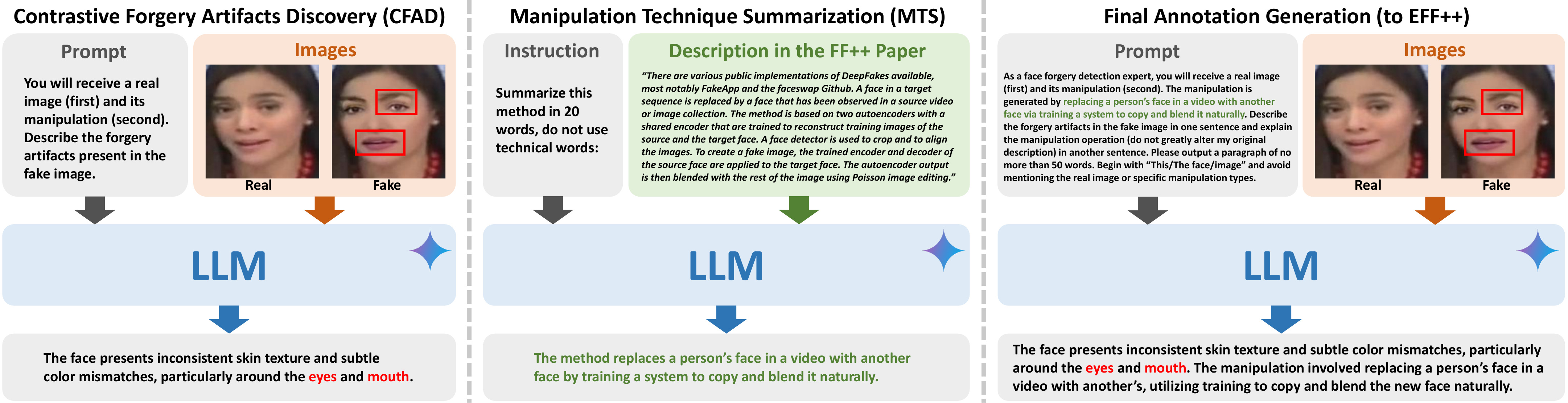}}
			\centering
		\end{minipage}
	\end{center}
	\caption{Annotation for a fake video frame from the EFF++ dataset. CFAD is employed to identify and explain facial forgeries, while MTS is used to extract and summarize the specific manipulation strategy applied. The combination of these methods results in an informative textual annotation. \label{prompt}}
\end{figure}

\section{Methodology}
\subsection{EFF++}
EFF++ is a frame-level, explainability-driven extension of the FF++ dataset \cite{Rossler_2019_ICCV}, developed using Gemini 2.0\footnote{\url{https://gemini.google.com}}, a commercial LLM with advanced multimodal alignment capabilities \cite{rahman2025comparative}. In EFF++, each manipulated video frame is accompanied by a textual annotation that describes both the forgery artifacts and the specific manipulation technique applied. As illustrated in Figure~\ref{prompt}, these annotations are generated through two novel methods: Contrastive Forgery Artifacts Discovery (CFAD), which identifies and explains facial forgeries, and Manipulation Technique Summarization (MTS), which concisely summarizes the manipulation strategies employed. The following section introduces CFAD and MTS and explains how they are integrated to produce the final textual annotations in EFF++.

\noindent\textbf{CFAD.}
Unlike previous methods that provide the LLM with only the manipulated image \cite{zhang2024common,chen2024textit}, we supply both the original and manipulated images to the LLM. Specifically, for each identity, we extract frames and crop faces from both the original video and its four manipulated counterparts, following the procedure detailed in Appendix~\ref{A11}. This yields a set of image pairs denoted as $\mathcal{S}=\{(I^i_R, I^{i,j}_F)\mid i=1,...,N;~j=1,...,4\}$, where $I_R$ and $I_F$ represent the real and fake images, respectively, and $N$ is the total number of frames. Subsequently, each image pair $(I_R^i, I_F^{i,j})$ is fed into the LLM, along with a prompt that clarifies the relationship between the two images and requests an explanation of any facial forgery clues present in the manipulated image. By enabling a direct visual comparison, CFAD empowers the LLM to more precisely identify and localize forgery artifacts.

\noindent\textbf{MTS.}
To enrich our dataset with more authoritative information beyond basic common-sense reasoning (e.g., \cite{zhang2024common, chen2024textit}), we incorporate brief explanations of the four FF++ manipulation techniques. Specifically, we extract the official descriptions of Deepfakes (DF), Face2Face (F2F) \cite{Thies_2016_CVPR}, FaceSwap (FS), and NeuralTextures (NT) \cite{thies2019deferred} from the FF++ paper. These are then provided to the LLM, which is instructed to generate clear, non-technical summaries. This process yields four one-sentence explanations, each capturing the essence of a specific manipulation strategy in language accessible to a broad audience. Both the original descriptions and the summaries are included in Appendix~\ref{A12}.

\noindent\textbf{Final Annotation Generation.}
In EFF++, each manipulated video frame is annotated using a combination of the CFAD and MTS methods. Specifically, we embed the MTS-generated summary (shown as green text in Figure~\ref{prompt}) within the CFAD's prompting framework. To further enhance the annotation process, we introduce supplemental instructions that (1) encourage subtle rephrasing of the summary description and (2) enforce a standardized output format. This approach ensures that each textual annotation captures both the forgery artifacts and the manipulation technique in a flexible yet consistent manner. For each real video frame, we employ a straightforward prompt that instructs the LLM to articulate the frame's natural, unmodified characteristics. Additional examples from the EFF++ dataset, as well as the prompt used for annotating real frames, are available in Appendix~\ref{A13}. Lastly, each annotation is prefixed with ``Yes'' (for manipulated video frames) or ``No'' (for real video frames), followed by the question, ``Is this image manipulated?'', thereby forming the EFF++ dataset.

\begin{figure}[t]
	\begin{center}
		\begin{minipage}{1\linewidth}
			{\includegraphics[width=1\linewidth]{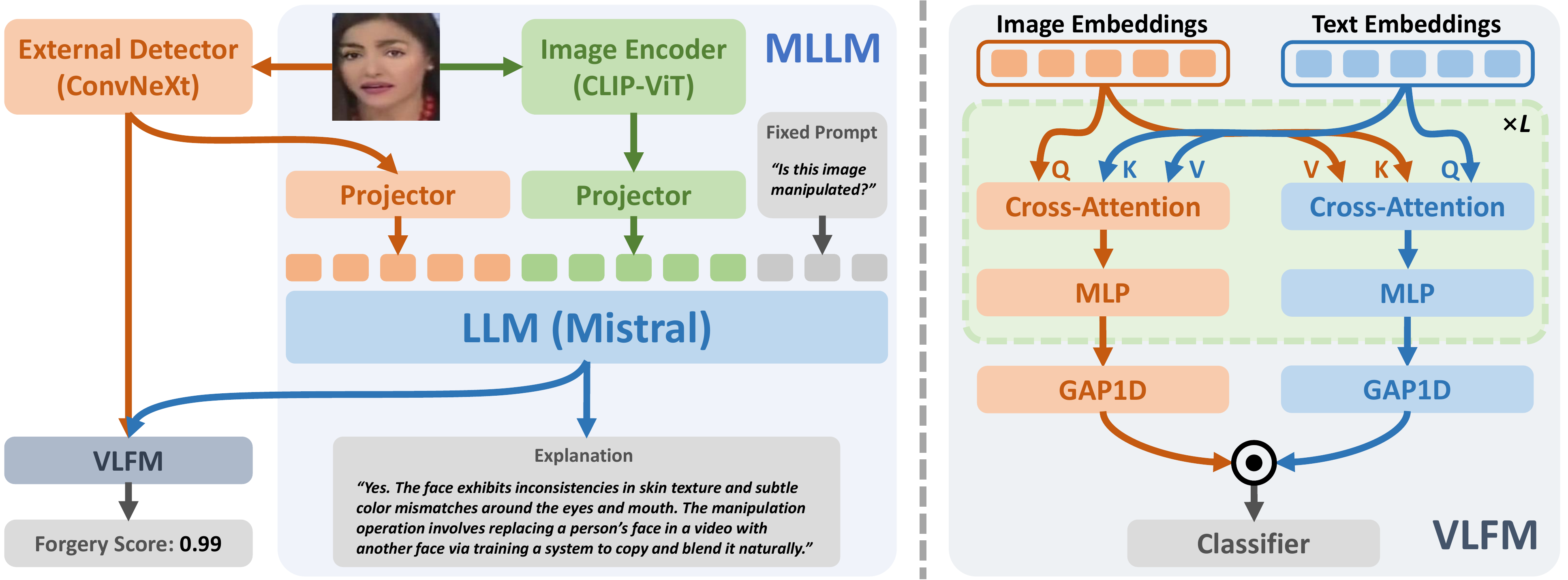}}
			\centering
		\end{minipage}
	\end{center}
	\caption{Overview of the VLF-Net architecture. The network consists of three main components: (1) an external detector, (2) an MLLM that includes an image encoder, two projectors, and an LLM, and (3) the innovative Vision-Language Fusion Module (VLFM). VLF-Net facilitates bidirectional interaction between visual and textual information. GAP1D refers to Global Average Pooling 1D. \label{vlfnet}}
\end{figure}

\subsection{VLF-Net}
\noindent\textbf{Overall Architecture.}
The VLF-Net, as illustrated in Figure~\ref{vlfnet}, is composed of three main components: an external detection, an MLLM, and the proposed Vision-Language Fusion Module (VLFM). For the external detector, we choose the widely used ConvNeXt architecture \cite{Liu_2022_CVPR}, which excels at capturing fine-grained visual features. The MLLM, built upon the LLaVA framework \cite{Liu_2024_CVPR}, includes an image encoder (CLIP-ViT \cite{pmlr-v139-radford21a}), two projectors, and an LLM (Mistral \cite{jiang2023mistral7b}). The VLFM, an innovative fusion structure, will be explained later. Given an input image, VLF-Net simultaneously processes it through the external detector and the image encoder, producing a visual feature map and ViT image embeddings, denoted as $E_{V}$. The visual feature map is then flattened across its spatial dimensions to generate ConvNeXt image embeddings, $E_{C}$. Subsequently, two distinct projectors map $E_{V}$ and $E_{C}$ into a text embedding space. These projected embeddings are concatenated with fixed prompt embeddings and passed to the LLM, which outputs text embeddings, $E_{L}$, as well as explanatory content. Finally, the VLFM integrates $E_{C}$ and $E_{L}$ to produce the classification result.

\noindent\textbf{VLFM.}
As depicted in Figure~\ref{vlfnet}, the VLFM receives image embeddings $E_C\in\mathbb{R}^{n_c\times d}$ and text embeddings $E_L\in\mathbb{R}^{n_l\times d}$ as inputs, where $n_c$ and $n_l$ denote the number of image and text tokens, respectively, and $d$ is the embedding dimension. To facilitate effective multimodal classification, VLFM employs a series of cross-modal fusion layers, each consisting of two complementary cross-attention mechanisms: an image-to-text block and a text-to-image block, along with corresponding Multi-Layer Perceptrons (MLPs). The image-to-text cross-attention block, $\mathrm{CA}_{\mathrm{img}\rightarrow \mathrm{text}}$, treats image embeddings $E_C$ as Queries (Q) and text embeddings $E_L$ as Keys (K) and Values (V), computed as:
\begin{equation}
	\mathrm{CA}_{\mathrm{img}\rightarrow \mathrm{text}}(E_C,E_L)=\mathrm{softmax}\left(\frac{E_CW^Q(E_LW^K)^\top}{\sqrt{d}}\right)(E_LW^V),
\end{equation}
where $W^Q,W^K,W^V\in\mathbb{R}^{d\times d}$ are learnable projection metrics. Conversely, the text-to-image cross-attention block, $\mathrm{CA}_{\mathrm{text}\rightarrow \mathrm{img}}$, uses text embeddings $E_L$ as Q and image embeddings $E_C$ as K and V:
\begin{equation}
	\mathrm{CA}_{\mathrm{text}\rightarrow \mathrm{img}}(E_L,E_C)=\mathrm{softmax}\left(\frac{E_LW^Q(E_CW^K)^\top}{\sqrt{d}}\right)(E_CW^V).
\end{equation}
Each cross-attention output is subsequently refined by an MLP. The VLFM comprises $L$ cross-modal fusion layers, progressively enhancing the integration of visual and textual features. Following the final fusion layer, the output embeddings are aggregated through Global Average Pooling 1D (GAP1D) and combined using a dot-product operation to produce a joint representation $Z$:
\begin{equation}
	Z = \mathrm{GAP1D}(E_C^{(L)}) \odot \mathrm{GAP1D}(E_L^{(L)}),
\end{equation}
where $E_C^{(L)}$ and $E_L^{(L)}$ represent the image and text embeddings after $L$ fusion layers, and $\odot$ denotes the dot-product. The joint representation $Z$ is then fed into a classifier to generate the final prediction.

\begin{figure}[t]
	\begin{center}
		\begin{minipage}{1\linewidth}
			{\includegraphics[width=1\linewidth]{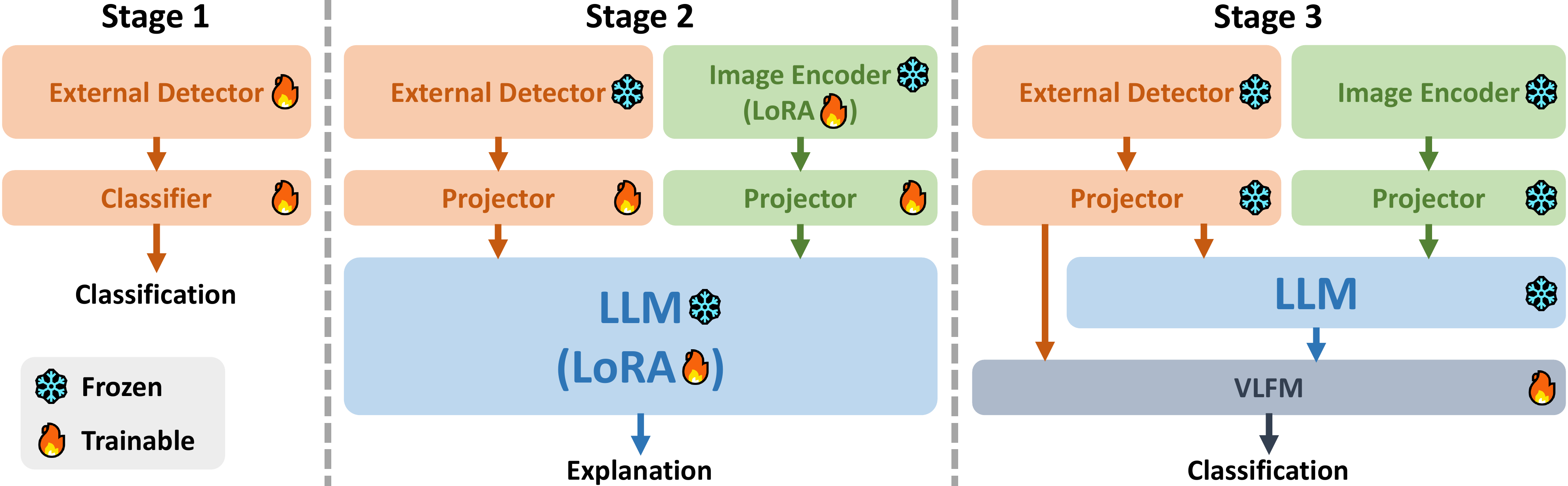}}
			\centering
		\end{minipage}
	\end{center}
	\caption{Overview of the three-stage training pipeline for VLF-Net. In Stage 1, the external detector is trained independently. In Stage 2, the external detector is frozen while the MLLM is fine-tuned. In Stage 3, both the external detector and the MLLM are frozen, and the VLFM is trained. \label{pipeline}}
\end{figure}

\noindent\textbf{Training Pipeline.}
We develop a three-stage training pipeline to maximize the performance of VLF-Net, as shown in Figure~\ref{pipeline}. In Stage 1, we train the external detector independently using standard Cross-Entropy (CE) Loss to extract visual features. In Stage 2, the external detector is frozen, and the MLLM is trained with Supervised Fine-Tuning (SFT) Loss to enhance textual representations. During this stage, both the image encoder and LLM are fine-tuned using Low-Rank Adaptation (LoRA) \cite{hu2022lora}, while the projectors undergo full training. Finally, in Stage 3, we freeze both the external detector and the MLLM, and train the VLFM with CE Loss to integrate visual and textual information.

\begin{table}[t]	
	\centering\renewcommand\arraystretch{1.2}\setlength{\tabcolsep}{7.8pt}
	\belowrulesep=0pt\aboverulesep=0pt
	\caption{Cross-dataset performance comparison of SOTA face forgery detection methods on the CDF2, DFDC, DFDCP, and FFIW benchmarks. Methods with $\ast$ are reproduced using their official codes, while results for others are taken from the original papers. The highest performance for each dataset is shown in \textbf{bold}, and the second-best is \underline{underlined}. Methods above the central dividing line are traditional approaches, while those below are MLLM-based. Notably, VLF-FFD consistently outperforms other methods across all datasets, demonstrating its superior generalization capability. \label{cross-data}}
	\begin{tabular}{c|c|c|cccc}
		\toprule
		\multirow{2}{*}{Method} & 
		\multirow{2}{*}{Venue} & 
		\multirow{2}{*}{Explainable} & 
		\multicolumn{4}{c}{Test Set AUC (\%)}\\
		\cmidrule(lr){4-7}
		&\multicolumn{1}{c|}{} 
		&\multicolumn{1}{c|}{} 
		&\multicolumn{1}{c}{CDF2} 
		&\multicolumn{1}{c}{DFDC} 
		&\multicolumn{1}{c}{DFDCP}
		&\multicolumn{1}{c}{FFIW} \\
		\midrule
		 F\textsuperscript{3}-Net\textsuperscript{$\ast$} \cite{qian2020thinking} & ECCV 2020 & \ding{55}  & 77.92 & 67.35 & 73.54 & 70.11 \\
		 LTW\textsuperscript{$\ast$} \cite{sun2021domain} & AAAI 2021 & \ding{55} & 77.14 & 69.00 & 74.58 & 76.63 \\
		 PCL+I2G \cite{Zhao_2021_ICCV} & ICCV 2021 & \ding{55} & 90.03 & 67.52 & 74.37 & - \\
		 DCL \cite{sun2022dual} & AAAI 2022 & \ding{55} & 82.30 & - & 76.71 & 71.14 \\
		 SBI \cite{Shiohara_2022_CVPR} & CVPR 2022 & \ding{55} & 93.18 & 72.42 & 86.15 & {84.83} \\
		 F\textsuperscript{2}Trans \cite{10004978} & TIFS 2023 & \ding{55} & 89.87 & - & 76.15 & - \\
		 AUNet \cite{Bai_2023_CVPR} & CVPR 2023 & \ding{55} & 92.77 & 73.82 & 86.16 & 81.45 \\
		 SeeABLE \cite{Larue_2023_ICCV} & ICCV 2023 & \ding{55} & 87.30 & 75.90 & 86.30 & - \\
		 LAA-Net \cite{Nguyen_2024_CVPR} & CVPR 2024 & \ding{55} & 95.40 & - & 86.94 & - \\
		 RAE \cite{10.1007/978-3-031-72943-0_23} & ECCV 2024 & \ding{55} &  \underline{95.50} & {80.20} & 89.50 & - \\
		 FreqBlender \cite{zhou2024freqblender} & NeurIPS 2024 & \ding{55}  & 94.59 & 74.59 & 87.56 & 86.14 \\
		 UDD \cite{fu2025exploring} & AAAI 2025 & \ding{55} & 93.10 & 81.20 & 88.10 & - \\
		 LESB \cite{Soltandoost_2025_WACV} & WACVW 2025 & \ding{55} & 93.13 & 71.98 & - & 83.01 \\
		\midrule
		 $\mathcal{X}^2$-DFD \cite{chen2024textit} & arXiv 2024 & \ding{51}  & \underline{95.50} & \underline{85.30} & 91.20 & - \\
		 KFD \cite{yu2025unlocking} & arXiv 2025 & \ding{51}  & 94.71 & 79.12 & \underline{91.81} & - \\
		 M2F2-Det \cite{guo2025rethinking} & arXiv 2025 & \ding{51}  & 95.10 & - & 87.80 & \underline{88.70} \\
		 \midrule
		 VLF-FFD (Ours) & - & \ding{51}  & \textbf{97.29} & \textbf{85.41} & \textbf{92.04} & \textbf{89.15} \\      
		\bottomrule
	\end{tabular}
\end{table}

\section{Experiments}
\subsection{Setup}
\noindent\textbf{Datasets.}
We conduct model training and intra-dataset evaluation on the widely recognized FaceForensics++ (FF++) benchmark \cite{Rossler_2019_ICCV}, which comprises 1,000 real face videos and 4,000 fake videos generated by four forgery techniques: DeepFakes (DF), Face2Face (F2F) \cite{Thies_2016_CVPR}, FaceSwap (FS), and NeuralTextures (NT) \cite{thies2019deferred}. Our three-stage training pipeline utilizes the FF++ dataset as follows: for cross-dataset assessment, we adhere to the SBI framework \cite{Shiohara_2022_CVPR} and employ only real FF++ samples during Stages 1 and 3; for intra-dataset assessment, both real and fake samples are used in these stages. In both cases, Stage 2 leverages the proposed EFF++ dataset, an explainability-driven extension of FF++, to enhance model interpretability. Cross-dataset performance is evaluated on four challenging datasets: Celeb-DeepFake-v2 (CDF2) \cite{Li_2020_CVPR}, which applies advanced deepfake methods to YouTube celebrity videos; the DeepFake Detection Challenge (DFDC) \cite{dolhansky2020deepfake} and its Preview version (DFDCP) \cite{dolhansky2019dee}, both of which introduce diverse perturbations such as compression, downsampling, and noise; and FFIW-10K (FFIW) \cite{Zhou_2021_CVPR}, which increases complexity by including multi-person scenarios.

\noindent\textbf{Baselines.}
We evaluate our approach against eighteen representative baselines for face forgery detection, including fifteen traditional methods and three MLLM-based techniques (introduced in Section~\ref{s22}). The traditional baselines encompass a diverse range of strategies: Face X-ray \cite{Li_2020_CVPR2} introduces a grayscale representation to reveal blend boundaries, while F\textsuperscript{3}-Net \cite{qian2020thinking} utilizes frequency-aware clues to identify manipulations. LTW \cite{sun2021domain} emphasizes domain-general robustness, and PCL+I2G \cite{Zhao_2021_ICCV} targets source-feature inconsistencies. DCL \cite{sun2022dual} exploits multi-granular contrastive learning for enhanced representation, whereas SBI \cite{Shiohara_2022_CVPR} augments training with synthetically generated fake faces. SOLA \cite{Fei_2022_CVPR} is designed to reduce local feature anomalies for improved generalization, and F\textsuperscript{2}Trans \cite{10004978} fuses spatial and frequency domain traces to capture forgery artifacts. AUNet \cite{Bai_2023_CVPR} investigates facial Action Unit (AU) regions for more nuanced analysis, and SeeABLE \cite{Larue_2023_ICCV} reframes forgery detection as an Out-Of-Distribution (OOD) challenge. LAA-Net \cite{Nguyen_2024_CVPR} introduces heatmap-guided self-consistency attention, while RAE \cite{10.1007/978-3-031-72943-0_23} focuses on recovering authentic facial appearances from perturbations. FreqBlender \cite{zhou2024freqblender} generates pseudo-fake faces by blending frequency knowledge, UDD \cite{fu2025exploring} mitigates position and content biases through a dual-branch architecture, and LESB \cite{Soltandoost_2025_WACV} employs a Local Feature Discovery (LFD) mechanism to create self-blended samples. Collectively, these baselines provide a comprehensive comparison framework for assessing the proposed method.

\begin{table}[t]	
	\centering\renewcommand\arraystretch{1.2}\setlength{\tabcolsep}{11pt}
	\belowrulesep=0pt\aboverulesep=0pt
	\caption{Intra-dataset performance comparison of SOTA face forgery detection methods. For clarity, only the top-performing results are highlighted. VLF-FFD achieves the best overall performance. \label{intra-data}}
	\begin{tabular}{c|c|cccc|c}
		\toprule
		\multirow{2}{*}{Method} & 
		\multirow{2}{*}{Venue} &
		\multicolumn{5}{c}{Test Set AUC (\%)} \\
		\cmidrule(lr){3-7}
		&\multicolumn{1}{c|}{} 
		&\multicolumn{1}{c}{DF} 
		&\multicolumn{1}{c}{F2F} 
		&\multicolumn{1}{c}{FS}
		&\multicolumn{1}{c|}{NT}
		&\multicolumn{1}{c}{FF++} \\
		\midrule
		Face X-ray\textsuperscript{$\ast$} \cite{Li_2020_CVPR2} & CVPR 2020 & 99.12 & 99.31 & 99.09 & 99.27 & 99.20 \\
		PCL+I2G \cite{Zhao_2021_ICCV} & ICCV 2021 & \textbf{100.00} & 99.57 & \textbf{100.00} & 99.58 & 99.79 \\
		SOLA \cite{Fei_2022_CVPR} & CVPR 2022 & \textbf{100.00} & 99.67 & \textbf{100.00} & 99.82 & 99.87 \\
		AUNet \cite{Bai_2023_CVPR} & CVPR 2023 & \textbf{100.00} & 99.86 & {99.98} & 99.71 & 99.89 \\
		RAE \cite{10.1007/978-3-031-72943-0_23} & ECCV 2024 & {99.60} & 99.10 & {99.20} & 97.60 & 98.90 \\
		\midrule
		VLF-FFD (Ours) & - & \textbf{100.00} & \textbf{99.89} & \textbf{100.00} & \textbf{99.73} & \textbf{99.91}\\      
		\bottomrule
	\end{tabular}
\end{table}

\begin{figure}[t]
	\begin{center}
		\begin{minipage}{1\linewidth}
			{\includegraphics[width=1\linewidth]{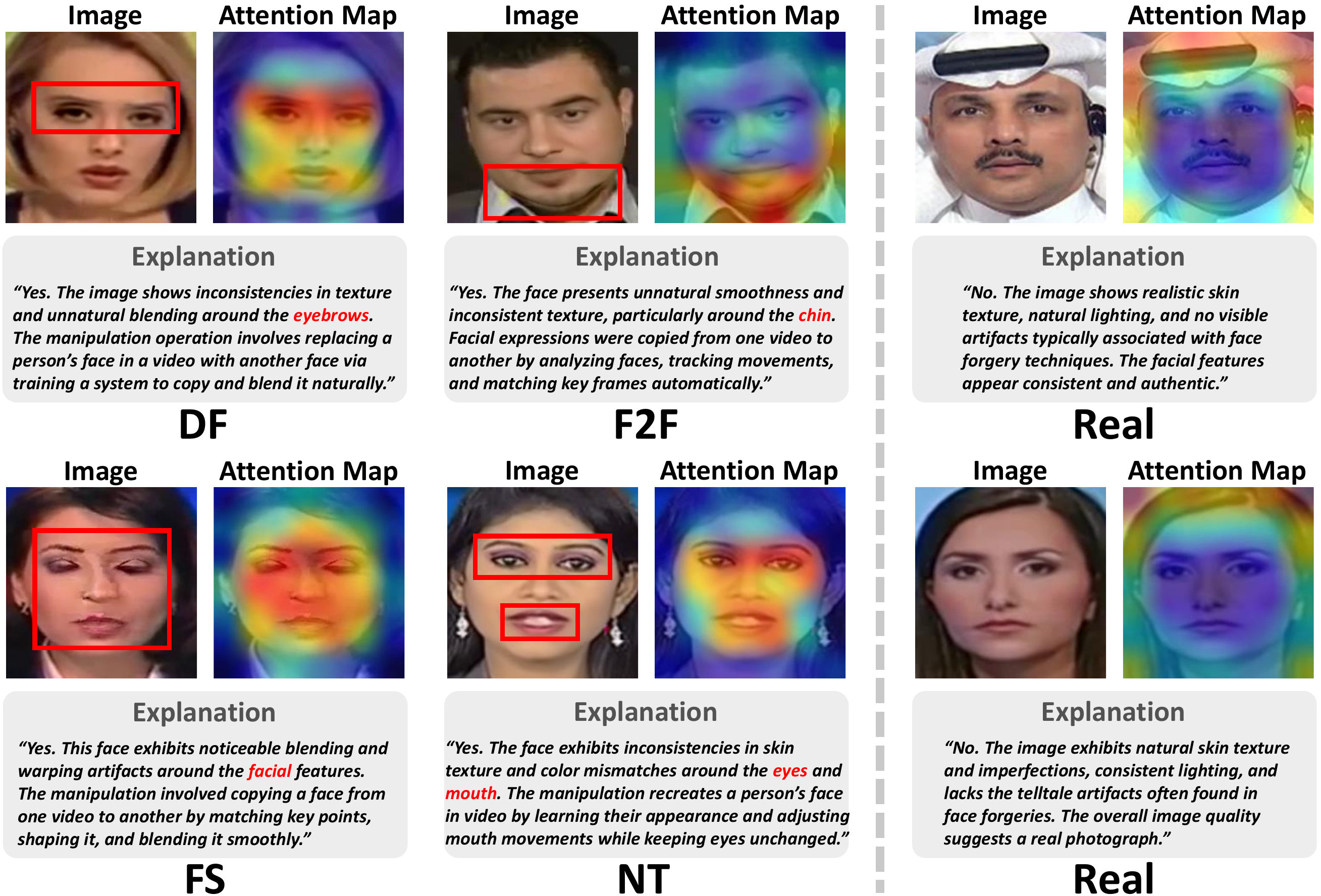}}
			\centering
		\end{minipage}
	\end{center}
	\caption{Attention maps generated by VLFM alongside their corresponding explanatory outputs for examples from the FF++ dataset. The attention maps highlight facial regions containing forgery artifacts, closely matching the areas mentioned in the explanatory contents. Notably, this visualization demonstrates the strong alignment between visual and textual modalities achieved by VLF-FFD. \label{attn}}
\end{figure}

\noindent\textbf{Evaluation Metric.}
Detection performance is evaluated using the Area Under the Receiver Operating Characteristic Curve (AUC), which is the standard metric for assessing face forgery detection. We report results at the video level by averaging predictions across all frames within each video.

\noindent\textbf{Implementation Details.}
This work leverages the preprocessing techniques, data augmentation strategies, and testing pipeline developed by SBI \cite{Shiohara_2022_CVPR}. For the external detector, we employ ConvNeXt-B-224 \cite{Liu_2022_CVPR} as the backbone, pre-trained on the ImageNet-1K dataset \cite{5206848}. The MLLM is based on LLaVA-v1.6-Mistral-7B \cite{Liu_2024_CVPR}, which utilizes CLIP-ViT-L/14-336 \cite{pmlr-v139-radford21a} as the image encoder and Mistral-7B \cite{jiang2023mistral7b} as the LLM. Additionally, the number of cross-modal fusion layers $L$ is set to 2. The three-stage training pipeline is configured as follows: In Stage 1, the external detector is trained for 200 epochs using the AdamW optimizer \cite{loshchilov2018decoupled}, a batch size of 64, and an initial learning rate of $5\times 10^{-5}$, with linear learning rate decay commencing at epoch 100 to encourage stable convergence. Stage 2 involves training the MLLM for 6 epochs with a batch size of 32. In Stage 3, we train the VLFM for 5 epochs with a batch size of 128, again employing the AdamW optimizer and a learning rate of $5\times 10^{-5}$. All experiments are implemented using the PyTorch framework and executed on a cluster of eight RTX 3090 GPUs. Additional implementation details are provided in Appendix~\ref{B11}.

\begin{table}[t]
	\centering\renewcommand\arraystretch{1.2}
\begin{minipage}[t]{0.49\linewidth}
	\centering\setlength{\tabcolsep}{3.9pt}
	\scriptsize\belowrulesep=0pt\aboverulesep=0pt
	\caption{Ablation study results for the EFF++, showcasing its effectiveness in boosting detection performance compared to alternative datasets.}
	\label{EFF++}
	\begin{tabular}{c|cccc|c}
		\toprule
		\multirow{2}{*}{Dataset} & 
		\multicolumn{4}{c|}{Test Set AUC (\%)} &
		\multirow{2}{*}{Avg.} \\
		\cmidrule(lr){2-5}
		&\multicolumn{1}{c}{CDF2} 
		&\multicolumn{1}{c}{DFDC} 
		&\multicolumn{1}{c}{DFDCP}
		&\multicolumn{1}{c|}{FFIW} \\
		\midrule
		DDVQA \cite{zhang2024common}  & 96.90 & 84.86 & 91.65 & 88.87 & 90.57 \\
		w/o CFAD \& MTS   & 97.12 & 85.11 & 91.70 & 88.99 & 90.73 \\
		w/o CFAD  & 97.09 & 85.03 & 91.76 & 89.14 & 90.76 \\
		w/o MTS & {97.25} & 85.32 & 91.85 & \textbf{89.21} & 90.91 \\
		\midrule
		EFF++ (Ours) & \textbf{97.29} & \textbf{85.41} & \textbf{92.04} & {89.15} & \textbf{90.97} \\ 
		\bottomrule
	\end{tabular}
\end{minipage}
\hfill
\begin{minipage}[t]{0.49\linewidth}	
	\centering\setlength{\tabcolsep}{2.3pt}
	\scriptsize\belowrulesep=0pt\aboverulesep=0pt
	\caption{Ablation study results for MLLM, highlighting the performance improvements obtained by incorporating MLLM into our framework.}
	\label{llava}
	\begin{tabular}{c|cccc|c}
		\toprule
		\multirow{2}{*}{Method} & 
		\multicolumn{4}{c|}{Test Set AUC (\%)} &
		\multirow{2}{*}{Avg.} \\
		\cmidrule(lr){2-5}
		&\multicolumn{1}{c}{CDF2} 
		&\multicolumn{1}{c}{DFDC} 
		&\multicolumn{1}{c}{DFDCP}
		&\multicolumn{1}{c|}{FFIW} \\
		\midrule
		Pre-trained LLaVA  & 51.32 & 56.54 & 56.03 & 55.70 & 54.90 \\
		Fine-tuned LLaVA  & 76.99 & 65.67 & 76.23 & 65.27 & 71.04 \\
		LLaVA in VLF-Net  & 90.40 & {77.76} & 81.10 & {80.56} & 82.46 \\
		w/o ConvNeXt-to-LLaVA & {96.41} & 84.40 & 90.68 & 88.09 & 89.90 \\
		\midrule
		VLF-FFD (Ours) & \textbf{97.29} & \textbf{85.41} & \textbf{92.04} & \textbf{89.15} & \textbf{90.97} \\ 
		\bottomrule
	\end{tabular}
\end{minipage}
\end{table}

\begin{table}[t]
\begin{minipage}[t]{0.49\linewidth}		
	\centering\renewcommand\arraystretch{1.2}\setlength{\tabcolsep}{3.4pt}
	\scriptsize\belowrulesep=0pt\aboverulesep=0pt
	\caption{Ablation study results for the external detector. Notably, CNN-based models outperform the Transformer-based ViT, and VLF-FFD improves the performance of all external detectors.}
	\label{others}
	\begin{tabular}{c|cccc|c}
		\toprule
		\multirow{2}{*}{Method} & 
		\multicolumn{4}{c|}{Test Set AUC (\%)} &
		\multirow{2}{*}{Avg.} \\
		\cmidrule(lr){2-5}
		&\multicolumn{1}{c}{CDF2} 
		&\multicolumn{1}{c}{DFDC} 
		&\multicolumn{1}{c}{DFDCP}
		&\multicolumn{1}{c|}{FFIW}\\
		\midrule
		EfficientNet-B4 \cite{Shiohara_2022_CVPR} & 93.18 & 72.42 & 86.15 & 84.83 & 84.15 \\
		+ VLF-FFD & 95.38 & {81.46} & 88.21 & 87.50 & 88.14 \\
		\midrule
		ViT-B/16 \cite{dosovitskiy2020image} & 93.64 & 79.66 & 85.73 & 83.92 & 85.74 \\
		+ VLF-FFD & 95.13 & {82.82} & 87.44 & 86.17 & 87.89 \\
		\midrule
		ConvNeXt-B \cite{Liu_2022_CVPR} & {96.08} & {84.26} & {90.20} & {87.67} & {89.55} \\ 
		+ VLF-FFD & \textbf{97.29} & \textbf{85.41} & \textbf{92.04} & \textbf{89.15} & \textbf{90.97} \\ 
		\bottomrule
	\end{tabular}
\end{minipage}
\hfill
\begin{minipage}[t]{0.49\linewidth}		
	\centering\renewcommand\arraystretch{1.2}\setlength{\tabcolsep}{3.5pt}
	\scriptsize\belowrulesep=0pt\aboverulesep=0pt
	\caption{Ablation study results for VLFM. The proposed architecture consistently surpasses all alternative variants, highlighting its superior ability to integrate visual and textual information.}
	\label{vlfm}
	\begin{tabular}{c|cccc|c}
		\toprule
		\multirow{2}{*}{Method} & 
		\multicolumn{4}{c|}{Test Set AUC (\%)} &
		\multirow{2}{*}{Avg.} \\
		\cmidrule(lr){2-5}
		&\multicolumn{1}{c}{CDF2} 
		&\multicolumn{1}{c}{DFDC} 
		&\multicolumn{1}{c}{DFDCP}
		&\multicolumn{1}{c|}{FFIW} \\
		\midrule
		w/o Cross-Attentions & 96.54 & 84.60 & 91.25 & 88.45 & 90.21 \\
		w/o $\mathrm{CA}_{\mathrm{img}\rightarrow \mathrm{text}}$ & 96.91 & 85.28 & 91.62 & 88.93 & 90.69 \\
		w/o $\mathrm{CA}_{\mathrm{text}\rightarrow \mathrm{img}}$ & 96.86 & 85.08 & 91.49 & 88.39 & 90.46 \\
		w/ Addition & 96.94 & 84.91 & 91.73 & 88.54 & 90.53 \\
		w/ Concatenation & 97.07 & 85.00 & {91.81} & 88.84 & 90.68 \\ 
		\midrule
		VLFM (Ours) & \textbf{97.29} & \textbf{85.41} & \textbf{92.04} & \textbf{89.15} & \textbf{90.97} \\ 
		\bottomrule
	\end{tabular}
\end{minipage}
\end{table}

\subsection{Results}
\noindent\textbf{Cross-Dataset Evaluation.}
Table~\ref{cross-data} compares our approach with recent SOTA face forgery detectors across multiple unseen datasets. The proposed VLF-FFD outperforms both traditional and MLLM-based methods, demonstrating its exceptional generalization capability. This superior performance can be attributed to the effective integration of visual and textual information within our framework.

\noindent\textbf{Intra-Dataset Evaluation.}
Table~\ref{intra-data} presents the intra-dataset performance comparison between VLF-FFD and existing methods, underscoring the superior detection ability of our approach. Figure~\ref{attn} illustrates the attention maps produced by VLFM alongside their corresponding explanatory outputs. Notably, the attention maps consistently focus on facial regions exhibiting forgery artifacts, which closely align with the areas described in the generated explanations. This strong correspondence demonstrates the effective integration of visual and textual modalities achieved by VLF-FFD.

\begin{figure}[t]
	\begin{center}
		\begin{minipage}[t]{1\linewidth}
			\begin{minipage}[t]{0.49\linewidth}
				{\includegraphics[width=1\linewidth]{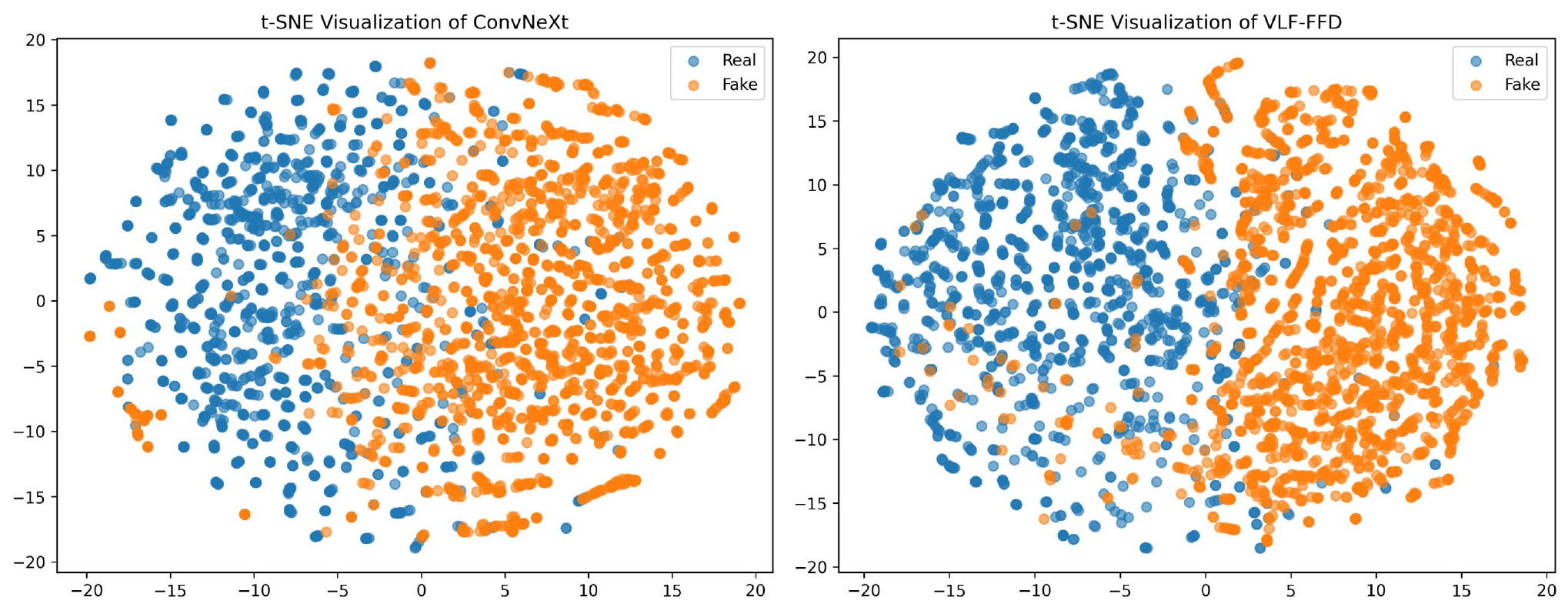}}
				{t-SNE results on CDF2}
				\centering
				
			\end{minipage}
			\begin{minipage}[t]{0.49\linewidth}
				{\includegraphics[width=1\linewidth]{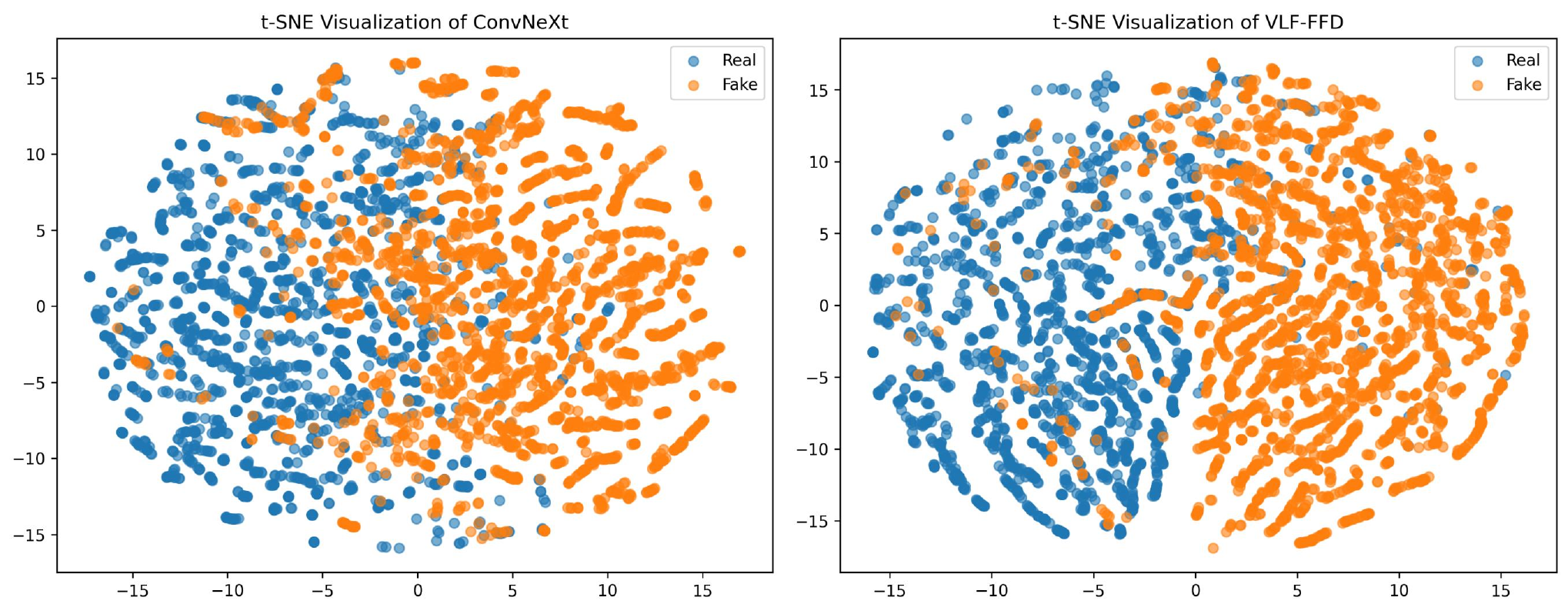}}
				{t-SNE results on DFDCP}
				\centering
				
			\end{minipage}
		\end{minipage}
	\end{center}
	\caption{t-SNE visualizations of feature representations produced by ConvNeXt and VLF-FFD models on the CDF2 and DFDCP datasets. The proposed VLF-FFD yields more distinct and well-separated clusters, highlighting its superior discriminative capability compared to ConvNeXt. \label{tsne}}
\end{figure}

\subsection{Ablation Studies}
In this section, we systematically discuss key components of our work: the proposed EFF++ dataset, our approach to integrating the MLLM within the VLF-FFD framework, our choice of external detectors, and the specific architectural designs of the VLFM. Additional ablation studies, such as investigations into LLM selection and the three-stage training strategy, are provided in Appendix~\ref{B12}.

\noindent\textbf{EFF++.}
To evaluate the impact of the EFF++ dataset on forgery detection performance, we conduct experiments by training VLF-Net on four distinct datasets during the second stage of our training pipeline. These datasets include: (1) DDVQA \cite{zhang2024common}, a video-level, explainability-focused extension of FF++; (2) a frame-level dataset generated using only the basic prompt (w/o CFAD \& MTS); (3) an EFF++ variant created without the CFAD method (w/o CFAD); and (4) an EFF++ variant produced without the MTS method (w/o MTS). As summarized in Table~\ref{EFF++}, the EFF++ dataset achieves the highest overall performance, validating our design choices. Specifically, frame-level annotations help mitigate overfitting, CFAD enables precise identification and localization of forgery artifacts, and MTS enriches explanations with authoritative information that goes beyond common-sense reasoning.

\noindent\textbf{MLLM.}
We conduct a series of ablation studies to quantify the advantages of embedding the MLLM within our VLF-FFD framework, rather than using it as a standalone forgery detector. Specifically, we compare our method against four variants: (1) the original LLaVA model without any fine-tuning (Pre-trained LLaVA); (2) the LLaVA model fine-tuned on the EFF++ dataset (Fine-tuned LLaVA); (3) the LLaVA model in VLF-Net, where prediction scores are derived from its textual explanations; and (4) VLF-Net without image embeddings $E_C$ fed into the LLM, effectively removing the connection between ConvNeXt and LLaVA. As Table \ref{llava} demonstrates, our fully integrated approach consistently outperforms all alternatives under cross-dataset evaluation. The notably poor performance of both pre-trained and fine-tuned LLaVA models underscores the need for specialized datasets and highlights the crucial role of external detectors in achieving robust MLLM-based face forgery detection.

\noindent\textbf{External Detector.}
We perform a series of experiments to assess the impact of various external detectors on VLF‑FFD's performance. Specifically, we compare ConvNeXt \cite{Liu_2022_CVPR} with another CNN-based model, EfficientNet \cite{Shiohara_2022_CVPR}, as well as the Transformer-based ViT \cite{dosovitskiy2020image}. We also evaluate the overall performance when each of these architectures is integrated into the VLF-FFD framework. As shown in Table~\ref{others}, the cross-dataset results highlight two key findings: (1) CNN-based backbones consistently outperform the Transformer‑based ViT within our framework, and (2) VLF-FFD substantially enhances the forgery detection capabilities of all external detectors tested. Additionally, t-SNE visualizations \cite{van2008visualizing} in Figure~\ref{tsne} demonstrate that VLF-FFD achieves a clearer separation between real and fake samples than standalone external detectors, highlighting its superior discriminative power.

\noindent\textbf{VLFM.}
To rigorously assess the effectiveness of VLFM, we conduct comprehensive ablation studies across multiple structural variants. Specifically, we analyze: (1) the overall contribution of cross-attention blocks by removing them entirely (w/o Cross-Attentions); (2) the individual roles of each cross-attention direction by ablating either $\mathrm{CA}_{\mathrm{img}\rightarrow \mathrm{text}}$ or $\mathrm{CA}_{\mathrm{text}\rightarrow \mathrm{img}}$; and (3) the relative advantages of the dot-product feature fusion method compared to alternative strategies, such as addition and concatenation. As shown in Table~\ref{vlfm}, the cross-dataset evaluations validate the design choices of VLFM, consistently demonstrating their substantial benefits to overall model performance.

\section{Conclusion}
This paper introduces VLF-FFD, a novel Vision-Language Fusion framework tailored for MLLM-enhanced Face Forgery Detection. Our contributions are twofold. First, we develop EFF++, an explainability-focused extension of the FF++ dataset that incorporates CFAD and MTS methods to generate precise and authoritative textual annotations at the frame level. Second, we propose VLF-Net, an innovative network architecture that facilitates bidirectional interaction between visual and textual modalities. To maximize the effectiveness of VLF-Net, we devise a structured three-stage training pipeline. Through extensive cross-dataset and intra-dataset experiments across five widely adopted benchmarks: CDF2, DFDC, DFDCP, FFIW, and FF++, we demonstrate that VLF-FFD consistently surpasses both traditional approaches and MLLM-based methods. Comprehensive ablation studies further validate the significance of each component within our framework. Collectively, these results establish VLF-FFD as an effective and robust solution for MLLM-driven face forgery detection.

%
%

\bibliographystyle{unsrt}  
\bibliography{references}


\appendix

\begin{figure}[h]
	\begin{center}
		\begin{minipage}{1\linewidth}
			{\includegraphics[width=1\linewidth]{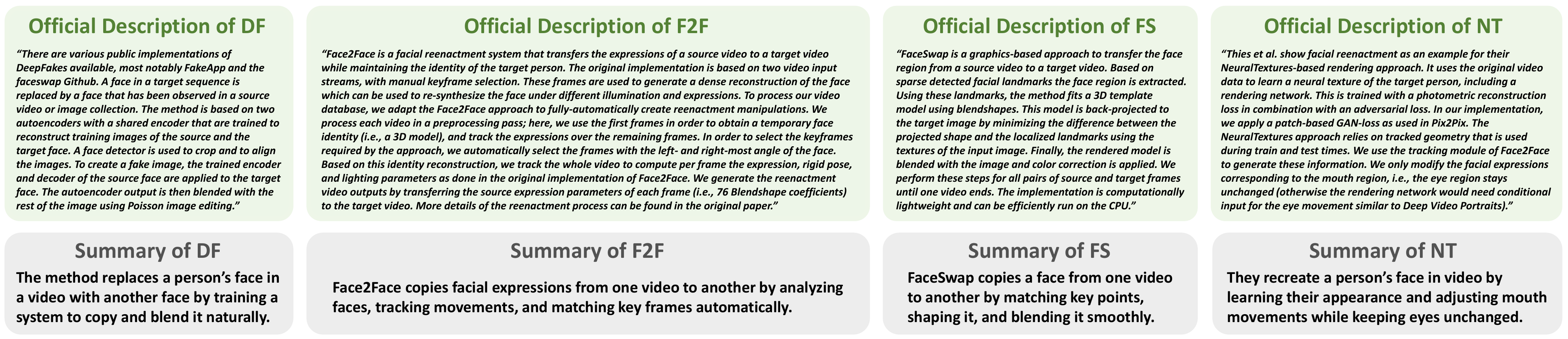}}
			\centering
		\end{minipage}
	\end{center}
	\caption{Official descriptions of the manipulation techniques and their MTS-generated summaries. \label{mts}}
\end{figure}

\section{Supplementary Information on EFF++}
\subsection{CFAD} 
\label{A11}
For each identity, we extract frames and crop faces from both the real video and its four fake versions, using identical settings to ensure consistency for the CFAD method. Because the real and fake videos may differ in length, we first determine the minimum frame count among the five videos. We then evenly sample frames from the start-up to this minimum frame index in each video, establishing a clear correspondence between real and fake sequences. For face cropping, we apply the strategy outlined in Appendix~\ref{B11}, ensuring uniformity between each real video frame and its fake counterparts.

\subsection{MTS}
\label{A12}
Figure~\ref{mts} presents the official descriptions of the four manipulation techniques used in the FF++ dataset, together with their corresponding summaries produced by MTS. Notably, the generated summaries are clear, concise, and non-technical, making them accessible to a wide audience.

\subsection{Final Annotation Generation} 
\label{A13}
\begin{figure}[t]
	\begin{center}
		\begin{minipage}{1\linewidth}
			{\includegraphics[width=1\linewidth]{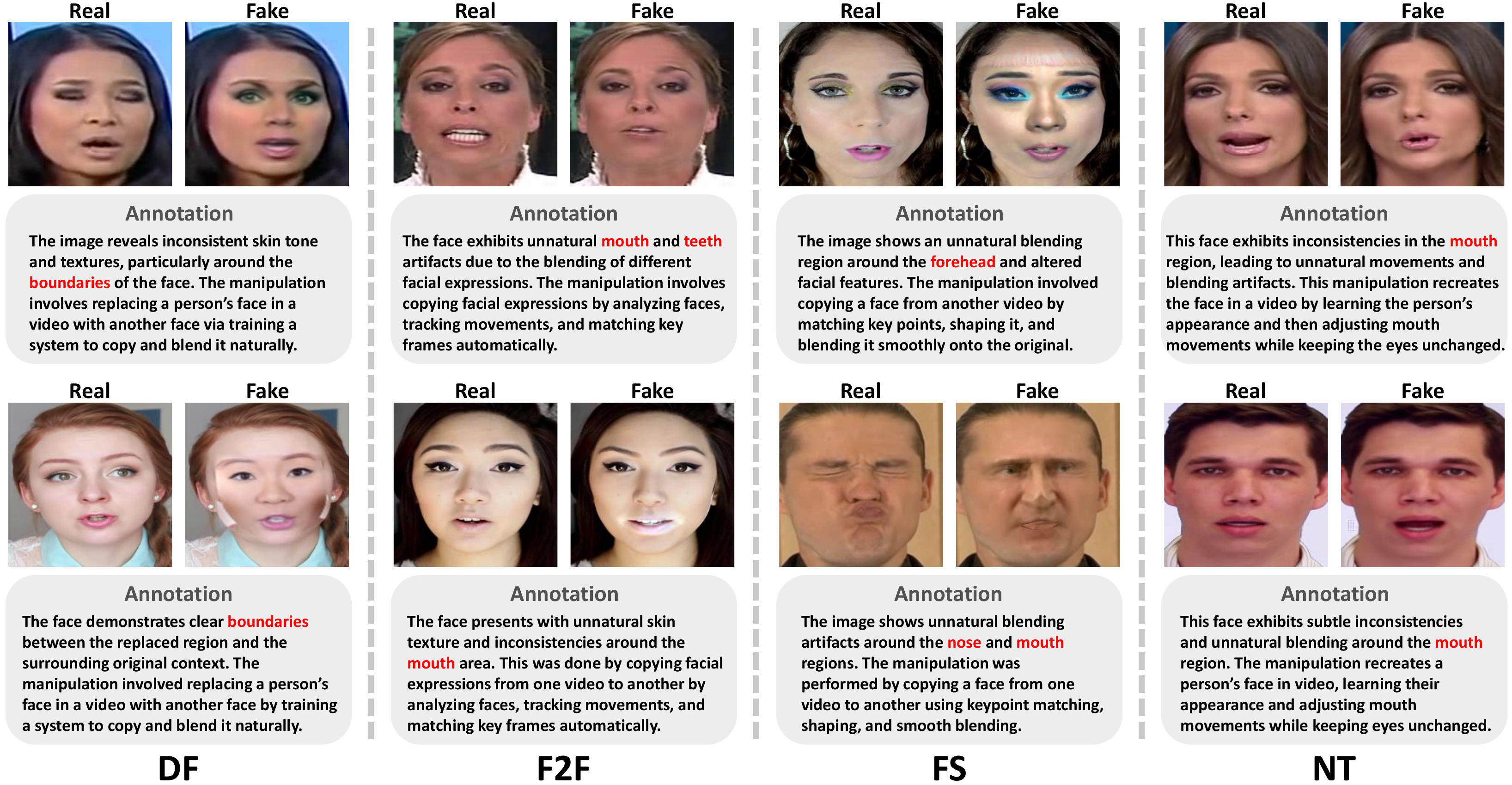}}
			\centering
		\end{minipage}
	\end{center}
	\caption{Example annotations of fake samples from the EFF++ dataset. Each textual annotation precisely identifies the forgery artifacts and explains the specific manipulation technique applied. \label{samples}}
\end{figure}

Figure~\ref{samples} presents several annotation examples from the EFF++ dataset. The textual annotations accurately describe the forgery artifacts present in fake images and provide authoritative explanations of the manipulation techniques used. For real video frames, we use the following prompt to produce textual annotations: \emph{``As a face forgery detection expert, you will receive a real image. Please provide a concise explanation of why it is real in no more than 40 words. Begin with `This/The image/face'.''}

\section{Supplementary Information on Experiments}
\subsection{Implementation Details}
\label{B11}
\noindent\textbf{SBI.} 
The SBI framework \cite{Shiohara_2022_CVPR} synthesizes fake face images from real samples by emulating the deepfake generation process. First, a real image is processed by two components: the Source-Target Generator (STG) and the Mask Generator (MG). The STG applies a series of image transformations to create pseudo source and target images, while the MG produces a blending mask based on pre-detected facial landmarks, which is then augmented to enhance diversity. Finally, the pseudo source and target images are blended using the generated mask, resulting in a pseudo-fake face sample.

\noindent\textbf{Preprocessing.} 
The 81-point facial landmark predictor from Dlib \cite{king2009dlib} is used to extract facial landmarks, which are required only during the training phase. For face detection, RetinaFace \cite{Deng_2020_CVPR} is employed to generate facial bounding boxes. During training, the bounding box is cropped with a random margin between 4\% and 20\%, whereas a fixed margin of 12.5\% is applied during inference. Additionally, for face cropping in CFAD, we employ a random margin between 4\% and 20\%.

\noindent\textbf{Data Augmentation.}
The image processing toolbox introduced in \cite{buslaev2020albumentations} is utilized for data augmentation. Within the STG, transformations such as RGBShift, HueSaturationValue, RandomBrightnessContrast, Downscale, and Sharpen are applied to generate pseudo source and target images. Furthermore, for real samples during training under the SBI framework, as well as for all samples in other training scenarios, augmentations such as ImageCompression, RGBShift, HueSaturationValue, and RandomBrightnessContrast are employed. These augmentations expose the model to a wider range of image variations, thereby improving its ability to generalize to previously unseen data.

\noindent\textbf{Settings for Training Stage 2.}
In the second training stage, we freeze the external detector and fine-tune the MLLM. Specifically, we set the rank and alpha parameters for LoRA tuning \cite{hu2022lora} to 128 and 256, respectively. The initial learning rate for both the projectors and LoRA layers is set to $1 \times 10^{-5}$ and is decayed to zero via a cosine schedule starting from the first iteration. The entire MLLM is trained using bfloat16 precision, with gradient accumulation performed after each iteration.

\begin{table}[t]
	\centering\renewcommand\arraystretch{1.2}
	\begin{minipage}[t]{0.49\linewidth}
		\centering\setlength{\tabcolsep}{1.4pt}
		\footnotesize\belowrulesep=0pt\aboverulesep=0pt
		\caption{Ablation study results for the LLM, validating the correctness of our design choice.}
		\label{llm}
		\begin{tabular}{c|cccc|c}
			\toprule
			\multirow{2}{*}{Method} & 
			\multicolumn{4}{c|}{Test Set AUC (\%)} &
			\multirow{2}{*}{Avg.} \\
			\cmidrule(lr){2-5}
			&\multicolumn{1}{c}{CDF2} 
			&\multicolumn{1}{c}{DFDC} 
			&\multicolumn{1}{c}{DFDCP}
			&\multicolumn{1}{c|}{FFIW} \\
			\midrule
			Vicuna-7B \cite{vicuna2023} & 96.70 & 84.95 & 91.38 & 88.23 & 90.32 \\
			Llama-3-8B \cite{grattafiori2024llama} & 97.22 & \textbf{85.46} & 91.90 & {88.69} & 90.82 \\
			\midrule
			Mistral-7B \cite{jiang2023mistral7b} & \textbf{97.29} & 85.41 & \textbf{92.04} & \textbf{89.15} & \textbf{90.97} \\ 
			\bottomrule
		\end{tabular}
	\end{minipage}
	\hfill
	\begin{minipage}[t]{0.49\linewidth}	
		\centering\setlength{\tabcolsep}{2.7pt}
		\footnotesize\belowrulesep=0pt\aboverulesep=0pt
		\caption{Ablation study results for training strategy, demonstrating the superiority of our method.}
		\label{train}
		\begin{tabular}{c|cccc|c}
			\toprule
			\multirow{2}{*}{Method} & 
			\multicolumn{4}{c|}{Test Set AUC (\%)} &
			\multirow{2}{*}{Avg.} \\
			\cmidrule(lr){2-5}
			&\multicolumn{1}{c}{CDF2} 
			&\multicolumn{1}{c}{DFDC} 
			&\multicolumn{1}{c}{DFDCP}
			&\multicolumn{1}{c|}{FFIW} \\
			\midrule
			One-Stage & 84.08 & 72.50 & 81.62 & 71.49 & 77.42 \\
			Two-Stage & 95.91 & 83.87 & 90.47 & 87.58 & 89.46 \\
			\midrule
			Three-Stage & \textbf{97.29} & \textbf{85.41} & \textbf{92.04} & \textbf{89.15} & \textbf{90.97} \\ 
			\bottomrule
		\end{tabular}
	\end{minipage}
\end{table}

\noindent\textbf{Other Details.} 
During training under the SBI framework, we sample eight frames from each real video. For other training scenarios, we sample 32 frames from each real video and eight frames from each fake video to maintain a balanced ratio of positive and negative samples. During testing, we sample 32 frames per video. If a frame contains multiple detected faces, we apply the classification model to each face individually and use the highest fakeness score as the predicted confidence for that frame. Within VLF-Net, the visual feature map from the penultimate stage of the ConvNeXt network is sent to the MLLM, while the feature map from the final stage is passed to the VLFM. Additionally, we utilize multi-head cross-attention blocks in the VLFM, with the number of heads set to 16.

\begin{table}[t]
	\centering\renewcommand\arraystretch{1.2}
	\centering\setlength{\tabcolsep}{12pt}
	\belowrulesep=0pt\aboverulesep=0pt
	\caption{Ablation study results on the number of cross-modal fusion layers in VLFM.}
	\label{hyper-para}
	\begin{tabular}{c|cccc|c}
		\toprule
		\multirow{2}{*}{Fusion Layers ($L$)} & 
		\multicolumn{4}{c|}{Test Set AUC (\%)} &
		\multirow{2}{*}{Avg.} \\
		\cmidrule(lr){2-5}
		&\multicolumn{1}{c}{CDF2} 
		&\multicolumn{1}{c}{DFDC} 
		&\multicolumn{1}{c}{DFDCP}
		&\multicolumn{1}{c|}{FFIW} \\
		\midrule
		1 & 97.13 & 85.24 & 91.53 & 89.01 & 90.73 \\
		2 & \textbf{97.29} & \textbf{85.41} & \textbf{92.04} & {89.15} & \textbf{90.97} \\ 
		3 & 97.03 & 85.36 & 91.52 & \textbf{89.46} & 90.84 \\
		4 & 97.01 & 84.80 & 91.42 & 88.96 & 90.55 \\
		\bottomrule
	\end{tabular}
\end{table}

\subsection{Ablation Studies}
\label{B12}
\noindent\textbf{LLM.} 
We compare our selected LLM, Mistral-7B \cite{jiang2023mistral7b}, with two popular models: Vicuna-7B \cite{vicuna2023} and Llama-3-8B \cite{grattafiori2024llama}. The cross-dataset evaluations, shown in Table~\ref{llm}, validate our design choice.

\noindent\textbf{Training Strategy.} 
We evaluate our three-stage training pipeline against two alternative strategies: (1) one-stage training, in which the external detector, MLLM, and VLFM are trained simultaneously, and (2) two-stage training, where the external detector is first optimized using the SBI framework, followed by joint training of the remaining VLF-Net components. As summarized in Table~\ref{train}, our three-stage training strategy achieves significantly better performance than the other two approaches, clearly demonstrating its superior effectiveness in maximizing the potential of VLF-Net.

\noindent\textbf{VLFM.} 
We evaluate the impact of varying the number of cross-modal fusion layers in VLFM. The cross-dataset results in Table~\ref{hyper-para} indicate that using two fusion layers yields optimal performance.

\section{Discussions}
\subsection{Strengths}
This paper exhibits two main advantages. First, the EFF++ dataset provides frame-level textual annotations that not only precisely describe forgery artifacts but also offer authoritative explanations of the manipulation techniques used. Second, the VLF-Net enables rich bidirectional interaction between visual and textual modalities, leading to SOTA performance in face forgery detection.

\subsection{Limitation}
This work focuses on frame-level face forgery detection. Therefore, the proposed framework may not be readily applicable to video-level face forgery detection without further adaptation.

\end{document}